\documentclass[11pt, oneside]{article}
\usepackage[margin=1in]{geometry}
\usepackage{float}
\usepackage{authblk}
\usepackage{amssymb, amsmath, amsthm}
\usepackage{bbold}

\usepackage[dvipsnames]{xcolor}
\definecolor{myblue}{RGB}{182,198,229}
\definecolor{myorange}{RGB}{247,203,175}
\usepackage{todonotes}
\usepackage{graphicx}
\usepackage{multirow}
\usepackage{hyperref}
\usepackage[backend=biber, style=numeric-comp, url=false,
doi=false, isbn=false, eprint=false, natbib=true,
sorting=none, uniquelist=false]{biblatex}
\AtEveryBibitem{\clearfield{month}}
\AtEveryBibitem{\clearfield{day}}
\AtEveryBibitem{\clearfield{version}}

\title{Generative weather for improved crop model simulations}
\author{Yuji Saikai\thanks{yuji.saikai@adelaide.edu.au}}
\affil{School of Agriculture, Food and Wine, The University of Adelaide}
\date{}
\begin{document}

\maketitle

\begin{abstract}
\noindent
Accurate and precise crop yield prediction is invaluable for decision making at both farm levels and regional levels. To make yield prediction, crop models are widely used for their capability to simulate hypothetical scenarios. While accuracy and precision of yield prediction critically depend on weather inputs to simulations, surprisingly little attention has been paid to preparing weather inputs. We propose a new method to construct generative models for long-term weather forecasts and ultimately improve crop yield prediction. We demonstrate use of the method in two representative scenarios---single-year production of wheat, barley and canola and three-year production using rotations of these crops. Results show significant improvement from the conventional method, measured in terms of mean and standard deviation of prediction errors. Our method outperformed the conventional method in every one of 18 metrics for the first scenario and in 29 out of 36 metrics for the second scenario. For individual crop modellers to start applying the method to their problems, technical details are carefully explained, and all the code, trained PyTorch models, APSIM simulation files and result data are made available.

\vspace{10pt}
\noindent
\textit{Keywords}: APSIM, fully convolutional networks, sequence modelling, time series, weather forecasts
\end{abstract}

\vspace{10pt}
\section{Introduction}\label{sec:intro}
Accurate and precise crop yield prediction is invaluable for decision making at both farm levels and regional levels. To make yield prediction, process-based crop models are widely used for their capability to simulate hypothetical scenarios. For example, farm-level decision support under variable weather conditions \cite{hochman_re-inventing_2009} and proactive management of regional food security risks \cite{fritz_comparison_2019}. Accuracy and precision of yield prediction critically depend on inputs to crop model simulations \cite{dokoohaki_comprehensive_2021}, in particular, weather inputs \cite{ramirez-villegas_assessing_2017}. However, despite the critical ingredient of crop models, little attention has been paid to weather input preparation.

For a crop model created to simulate some crop production system in a particular location over a specified number of days, the conventional approach to preparing daily weather inputs is simply to take subsequences of historical records at that location, where the length of each subsequence is the specified number of days. Examples are abundant in the literature, which is characterised by how many years of historical records are used: 20 years \cite{ojeda_effects_2020}, 30 years \cite{collins_improving_2021, tang_optimizing_2021, xiao_climate_2020}, 49 years \cite{dutta_improved_2020}, 62 years \cite{lilley_defining_2019} and location-specific years \cite{rahimi-moghaddam_towards_2021}. The conventional approach can be interpreted as implicitly modelling the future weather as uniformly distributed over a small set of subsequences of historical records and use random samples as inputs to crop models.

The lack of explicit modelling of future weather is understandable. Crop model simulations are often run for years, decades or even a century, implying that thousands of daily weather values must be prepared. Although preparation of these values may be regarded as weather forecasting, long-term weather forecasting is notoriously difficult due to the chaotic nature of the atmospheric system, as first conjectured by Lorenz \cite{lorenz_deterministic_1963}. Moreover, even if forecasts of acceptable quality can be produced by the operational numerical weather prediction (NWP), on-demand use of NWP for different locations over differing numbers of forecasting years will be prohibitively expensive \cite{bihlo_generative_2021,bauer_quiet_2015}.

To overcome the infeasibility of NWP and provide a practical method for weather input preparation, we propose neural network models that can be trained on historical records at locations of interest and generate desired numbers of future weather values. Over the past decade, there has been significant interest in machine learning for weather and climate modelling, signified by the theme issue on this topic \cite{philosophical_transactions_of_the_royal_society_a_machine_2021}. A key advantage of the machine learning approach over NWP is that models can be task-specific. In stark contrast to NWP, machine learning models can be flexibly designed for specific target variables (e.g., a subset of the atmospheric states for a specific location) and trained only on relevant datasets. Implications are the intrinsic absence of model bias and potential savings of computational resources \cite{schultz_can_2021}.

We realise this advantage for crop modellers by designing lightweight neural network architectures so that individual modellers can train unique neural networks and generate weather inputs to their crop models within reasonable amounts of time (e.g., 24 hours using desktops). For the purpose of generating weather inputs to crop models, we think that striking a balance between lightweight and capacity is crucial because many crop modellers have no access to expensive computing facilities used to train large neural networks (e.g., large language models). To this end, we adopt the fully convolutional network \cite{long_fully_2015}, which is simpler and cleaner than recurrent networks \cite{bai_empirical_2018} yet capable of generating complex sequence data \cite{gehring_convolutional_2017, van_den_oord_wavenet_2016}. Accordingly, we also avoid overly deep networks that typically require extra care such as residual connections \cite{he_deep_2016}.

In Section \ref{sec:problem}, we formally define the problem of weather input generation. In Section \ref{sec:likelihood}, we justify key restrictions needed for the balance in architectures. In Section \ref{sec:loss_fn}, we provide a detailed expression of the loss function. In Section \ref{sec:conv}, we explain dilated causal convolution, an innovation that enables lightweight sequence models. Following the explanation of the key concepts, in Section \ref{sec:architecture}, we explain our specific choice of neural network architectures. In Section \ref{sec:dataset}, we explain how modellers convert original time series datasets into specific training data. To demonstrate use of the method, we conduct experiments under two representative scenarios described in Section \ref{sec:experiments} with results reported in Section \ref{sec:results}. To address accuracy and precision of crop yield prediction, we calculate mean and standard deviation of absolute errors as performance metrics and compare them with the conventional method. Finally, in Section \ref{sec:discussion}, we discuss the results, merits and limitations of our method, and potential improvement for future research. Throughout the paper, we assume that crop models are created by the Agricultural Production Systems sIMulator (APSIM), one of the most popular crop modelling platforms. The method is readily applicable to other crop models by replacing the weather variables with suitable ones.

\section{Model}\label{sec:model}
\subsection{Weather generation problem}\label{sec:problem}
Let \(\{x_t\}_{t=1}^T\) be the daily time series of interest, where \(x_t, \forall t\in\{1,\dots,T\}\) is a vector consisting of 4 values required to run APSIM simulations \cite{apsim_initiative_creating_2024}:
\begin{itemize}
\item \(x_t^\mathrm{radn}\): solar radiation (MJ/m\(^2\))
\item \(x_t^\mathrm{mint}\): minimum temperature (°C)
\item \(x_t^\mathrm{maxt}\): maximum temperature (°C)
\item \(x_t^\mathrm{rain}\): rainfall (mm).
\end{itemize}
Two constraints on these variables are \(x_t^\mathrm{maxt}\geq x_t^\mathrm{mint}\) and \(x_t^\mathrm{radn}, x_t^\mathrm{rain}\geq 0\) at each \(t\). Let \(t_0\) denote the time step such that we generate samples of random variables \(\{x_t\}_{t=t_0+1}^T\) conditional on the observations of random variables \(\{x_t\}_{t=1}^{t_0}\). Formally, our goal is to estimate and take samples from the following distribution:
\begin{equation*}
p(x_{t_0+1},\dots,x_T|x_1,\dots,x_{t_0}).
\end{equation*}
In what follows, when it is clear from the context, we drop the conditioning part for notational ease and readability. For example, we may write \(x_t\) for \(x_t|x_1,\dots,x_{t-1}\)

\subsection{Likelihood}\label{sec:likelihood}
Our modelling strategy is based on the following decomposition:
\begin{equation}
p(x_{t_0+1},\dots,x_T|x_1,\dots,x_{t_0})
= \prod_{t=t_0+1}^T p(x_t|x_1,\dots,x_{t_0},\dots,x_{t-1}).\label{eq:target}
\end{equation}
Furthermore, within \(t\), \(p(x_t)\) is decomposed as follows:
\begin{align*}
p(x_t)
&= p(x_t^\mathrm{radn}, x_t^\mathrm{mint}, x_t^\mathrm{maxt}, x_t^\mathrm{rain})\\
&= p(x_t^\mathrm{radn})p(x_t^\mathrm{mint}|x_t^\mathrm{radn})p(x_t^\mathrm{maxt}|x_t^\mathrm{radn}, x_t^\mathrm{mint})p(x_t^\mathrm{rain}|x_t^\mathrm{radn}, x_t^\mathrm{mint}, x_t^\mathrm{maxt}).
\end{align*}
The specific order of conditional dependency incorporates the following intuition and modelling strategy. Since solar radiation is a cause of temperature, \(x_t^\mathrm{mint}\) and \(x_t^\mathrm{maxt}\) are conditioned on \(x_t^\mathrm{radn}\). In addition, since rainfall is the most erratic and challenging variable to predict, we try to facilitate its estimation by conditioning on the other variables.

Moreover, to incorporate the constraint \(x_t^\mathrm{mint}\leq x_t^\mathrm{maxt}\), we define a variable \(x_t^\mathrm{diff}=x_t^\mathrm{maxt}-x_t^\mathrm{mint}\) and preprocess the original data accordingly. Therefore, for all \(t\in\{t_0+1,\dots,T\}\), we estimate the following distribution
\begin{align}
\begin{split}    
p(x_t)
&= p(x_t^\mathrm{radn}, x_t^\mathrm{mint}, x_t^\mathrm{diff}, x_t^\mathrm{rain})\\
&= p(x_t^\mathrm{radn})p(x_t^\mathrm{mint}|x_t^\mathrm{radn})p(x_t^\mathrm{diff}|x_t^\mathrm{mint},x_t^\mathrm{radn})p(x_t^\mathrm{rain}|x_t^\mathrm{diff},x_t^\mathrm{mint},x_t^\mathrm{radn}).
\end{split}
\label{eq:likelihood}
\end{align}

As indicated by the growing size of the conditional part, \(x_1,\dots,x_{t_0},\dots,x_{t-1}\), as \(t\) increases from \(t_0+1\) to \(T\), we certainly do not assume distribution \(p(x_t)\) is identical for all \(t\in\{t_0+1,\dots,T\}\). However, to strike a balance between lightweight and capacity, we make the following distributional assumptions
\begin{align*}
x_t^\mathrm{radn} &\sim \mathrm{gamma}(\alpha_t^\mathrm{radn},\beta_t^\mathrm{radn})\\
x_t^\mathrm{mint}|x_t^\mathrm{radn} &\sim \mathrm{normal}(\mu_t^\mathrm{mint},\sigma_t^\mathrm{mint})\\
x_t^\mathrm{diff}|x_t^\mathrm{mint},x_t^\mathrm{radn} &\sim \mathrm{gamma}(\alpha_t^\mathrm{diff},\beta_t^\mathrm{diff})\\
x_t^\mathrm{rain}|x_t^\mathrm{diff},x_t^\mathrm{mint},x_t^\mathrm{radn} &\sim \mathrm{gamma}(\alpha_t^\mathrm{rain},\beta_t^\mathrm{rain}).
\end{align*}
The choice of distributions reflects that minimum temperature can be positive or negative, whereas the other variables must be non-negative. To this end, normal and gamma distributions are some of the simplest choices, and efficient computation of their densities is available in deep learning libraries including PyTorch, which we use for this paper.

\subsection{Loss function}\label{sec:loss_fn}
Given the likelihood specification, a natural choice for the loss function that is minimised to learn the neural network parameters is the negative log-likelihood. Let \(\theta\) be the neural network parameters to optimise; i.e., the parameters that specify all the convolution filters and biases. Then, the component negative log-likelihood for \(t\) on a single training example in the training data is formed by the specified probability density functions:
\begin{align}
\begin{split}    
l_{t}(\theta)
&= -\log\bigg(p(x_t^\mathrm{radn};\alpha_t^\mathrm{radn},\beta_t^\mathrm{radn})p(x_t^\mathrm{mint};\mu_t^\mathrm{mint},\sigma_t^\mathrm{mint}) \times\\
&\hspace{4.5em}p(x_t^\mathrm{diff};\alpha_t^\mathrm{diff},\beta_t^\mathrm{diff})p(x_t^\mathrm{rain};\alpha_t^\mathrm{rain},\beta_t^\mathrm{rain})\bigg)\\
&= -\log\bigg(
  \frac{(\beta^\mathrm{radn})^{\alpha^\mathrm{radn}}}{\Gamma(\alpha^\mathrm{radn})}(x_t^\mathrm{radn})^{\alpha^\mathrm{radn}-1}e^{-\beta x_t^\mathrm{radn}} \times\\
&\hspace{4.5em}\frac{1}{\sqrt{2\pi}\sigma_t^\mathrm{mint}}\exp\left(-\frac{1}{2}\left(\frac{x_t^\mathrm{mint}-\mu_t^\mathrm{mint}}{\sigma_t^\mathrm{mint}}\right)^2\right) \times\\
&\hspace{4.5em}\frac{(\beta^\mathrm{diff})^{\alpha^\mathrm{diff}}}{\Gamma(\alpha^\mathrm{diff})}(x_t^\mathrm{diff})^{\alpha^\mathrm{diff}-1}e^{-\beta x_t^\mathrm{diff}} \times\\
&\hspace{4.5em}\frac{(\beta^\mathrm{radn})^{\alpha^\mathrm{radn}}}{\Gamma(\alpha^\mathrm{radn})}(x_t^\mathrm{radn})^{\alpha^\mathrm{radn}-1}e^{-\beta x_t^\mathrm{radn}} \bigg)\\
&= -\alpha^\mathrm{radn}\log(\beta^\mathrm{radn})+\log(\Gamma(\alpha^\mathrm{radn}))-(\alpha^\mathrm{radn}-1)\log(x_t^\mathrm{radn})+\beta x_t^\mathrm{radn} +\\
&\hspace{1.5em}\frac{1}{2}\log(2\pi)+\log(\sigma_t^\mathrm{mint})+\frac{1}{2}\left(\frac{x_t^\mathrm{mint}-\mu_t^\mathrm{mint}}{\sigma_t^\mathrm{mint}}\right)^2 -\\
&\hspace{1.5em}\alpha^\mathrm{diff}\log(\beta^\mathrm{diff})+\log(\Gamma(\alpha^\mathrm{diff}))-(\alpha^\mathrm{diff}-1)\log(x_t^\mathrm{diff})+\beta x_t^\mathrm{diff} -\\
&\hspace{1.5em}\alpha^\mathrm{rain}\log(\beta^\mathrm{rain})+\log(\Gamma(\alpha^\mathrm{rain}))-(\alpha^\mathrm{rain}-1)\log(x_t^\mathrm{rain})+\beta x_t^\mathrm{rain}.
\end{split}
\label{eq:loss_fn}
\end{align}
where \(\Gamma\) is the gamma function. Here, \(x_t\) is the target (not the input) from a single training example, and \((\alpha_t^\mathrm{radn},\beta_t^\mathrm{radn})\), \((\mu_t^\mathrm{mint},\sigma_t^\mathrm{mint})\), \((\alpha_t^\mathrm{diff},\beta_t^\mathrm{diff})\) and \((\alpha_t^\mathrm{rain},\beta_t^\mathrm{rain})\) are the output of the neural network, hence a function of \(\theta\). The final loss function is the sum of \(l_{t}(\theta)\) over all \(t\in\{t_0+1,\dots,T\}\) and a batch of training examples used in computing the gradient. In Section \ref{sec:dataset}, we will describe how training data is constructed from the original time series dataset.

Note that all the likelihood parameters, except \(\mu_t^\mathrm{mint}\), must be positive. Thus, we use as the activation function after the last layer a combination of the identity function (for \(\mu_t\)) and a \(\mathrm{softplus}\) function with a small positive value (for the rest):
\[
\mathrm{softplus}(z) + \varepsilon = \log(1 + \exp(z)) + \varepsilon
\]
where \(\varepsilon=10^{-3}\) for numerical stability. We provide details of neural network architectures in Section \ref{sec:architecture}.

\subsection{Dilated causal convolution}\label{sec:conv}
\begin{figure}[H]
\centering
\includegraphics[width=1.\linewidth]{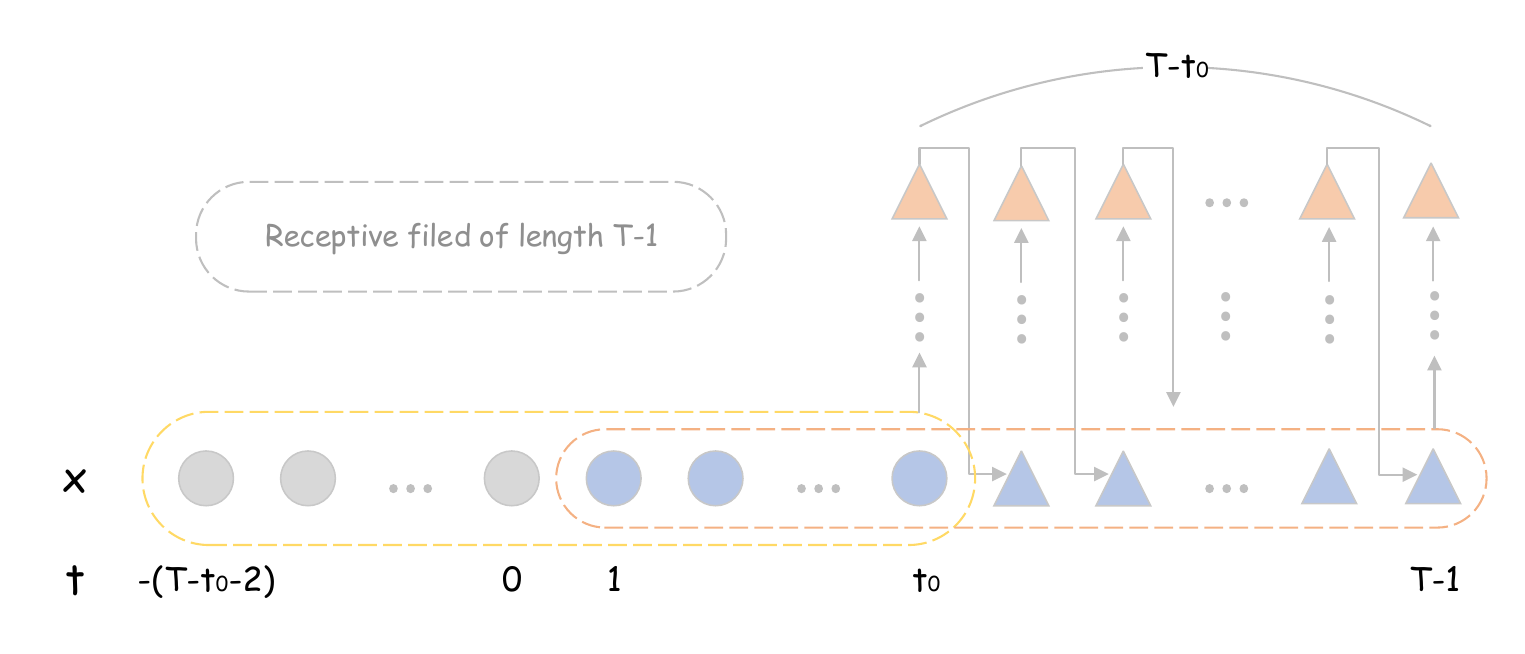}
\caption{\(\textcolor{myorange}{\blacktriangle}\) represents an output and \(\color{lightgray}\bullet\) represents a padded 0. Each output value is computed based on \(T-1\) values, which may consist of past observations \(\color{myblue}\bullet\), outputs from the preceding time steps \(\color{myblue}\blacktriangle\), and padded 0s \(\color{lightgray}\bullet\). Only two receptive fields, the one for the first output and the other for the last output are explicitly drawn. Notice that the last output is the only one whose receptive field does not contain any padded 0s.}
\label{fig:conv}
\end{figure}

In time series modelling, causality is intuitive and defined to be a restriction that a value at some time step is a function of values at only the preceding time steps. We may visualise 1-D convolution as sliding a filter over the input values placed on the timeline from left to right. When computing an output value at time \(t+1\), causality requires the rightmost slot of the filter must not go beyond the input time step \(t\). This causes inconsistency between causality and convolution because the former requires the growing number of input values as indicated by Equation \ref{eq:target}, whereas the latter operates on the fixed number of values equal to the filter length when computing each value for the output. A common solution is to use 0s as dummy input values, and we pad \(T-t_0-1\) number of 0s to the left of the original input \(\{x_t\}_{t=1}^{t_0}\) so that each output value is always computed using the same number of input values (with the decreasing number of padded 0s). We may think that the padded inputs have non-positive time indices: \(\{x_t\}_{t=-(T-t_0-2)}^{0}=0\). Figure \ref{fig:conv} illustrates key components and their relationships in causal convolution. In particular, notice that each output \(\textcolor{myorange}{\blacktriangle}\) is computed based on the same \(T-1\) number of inputs. This number is the size of the receptive field at the last layer, which is explained in the next paragraph.

\begin{figure}[H]
\centering
\includegraphics[width=\linewidth, clip=true, trim=.5cm .5cm .5cm .5cm]{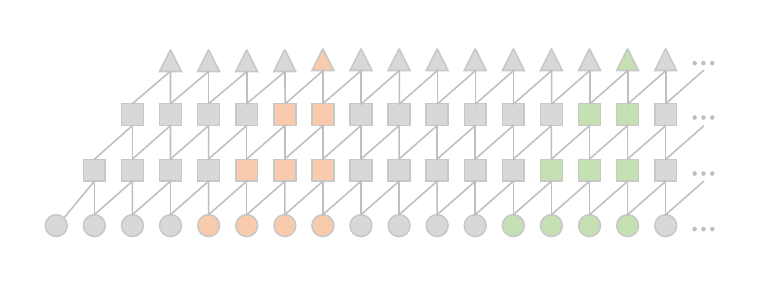}
\caption{Causal convolution using a filter of size 2 without dilation. The receptive field of \(\textcolor{myorange}{\blacktriangle}\) node is four \(\textcolor{myorange}{\bullet}\) at the input layer.}
\label{fig:causal}
\end{figure}

A receptive field defines which portion of the input is used to compute a particular part of the output of a convolutional layer. In 1-D convolution, the length of a filter used at a convolutional layer defines the receptive field. For example, in Figure \ref{fig:causal}, each output value at the first convolutional layer is computed based on two adjacent values (i.e., two adjacent circles in the figure) in the input layer. Hence, the receptive field of any node at the first convolutional layer is of length 2. Similarly, the receptive field of any node of the output of the second convolutional layer is of length 3, because each node is based on two nodes at the first layer (i.e., two adjacent squares in the figure), each of which is in turn based on two nodes at the input layer, and the middle input node contributes to both nodes at the first layer. As readers can imagine, to achieve a large receptive field at the final layer, we need large filters and/or many layers, both of which likely renders the training data-demanding and computationally expensive. For readability, in what follows, we use the term ``the receptive field`` to mean the receptive field of any node at the last layer, which is of our interest.

\begin{figure}[H]
\centering
\includegraphics[width=\linewidth, clip=true, trim=.5cm .5cm .5cm .5cm]{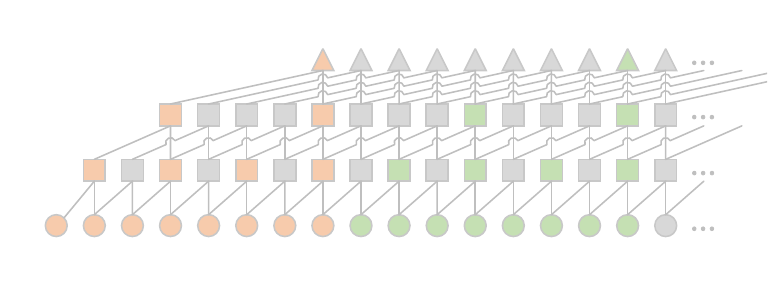}
\caption{Dilated causal convolution using a filter of size 2 with dilation factors of 1, 2, 4 and 8. The receptive field of \(\textcolor{myorange}{\blacktriangle}\) node is eight \(\textcolor{myorange}{\bullet}\) at the input layer. Without increasing the number of hidden layers or trainable parameters, dilated causal convolution achieves a larger receptive field.}
\label{fig:dilation}
\end{figure}

Use of dilated filters is one way to address the issue \cite{van_den_oord_wavenet_2016, borovykh_conditional_2018}. Based on a standard filter as a base filter, a dilated filter is constructed by inserting ``empty'' slots between two slots of the base filter. Empty slots do nothing but expanding the effective length of the filter without increasing the number of effective slots, which come with trainable parameters. For example, if the base filter is of length 2 and the dilation factor multiplied by the base length each time a convolutional layer is added, the effective length of the filter used at the fourth convolutional layer is \(9=2^{4-1}+1\) and, consequently, the length of the receptive field is \(16=2^4\) (Figure \ref{fig:dilation}).

\subsection{Neural network architecture}\label{sec:architecture}
Using dilated filters, we design neural network architectures to achieve the goal, estimating and sampling from Equation \ref{eq:target}. To sequentially output pairs of parameters \((\alpha_t,\beta_t)\) that specify the likelihood within each \(t\), input \(x\) is first processed by masked convolution. Specifically, \(x\) of size \(4\times t_0\) is processed by a 2-D filter of size \(4\times2\), where the second column of the filter is masked by:
\begin{itemize}
    \item \([0,0,0,0]\) to output for \(x_t^\mathrm{radn}\) based on \(x_1,\dots,x_{t-1}\)
    \item \([1,0,0,0]\) to output for \(x_t^\mathrm{mint}\) based on \(x_1,\dots,x_{t-1}\) and \(x_t^\mathrm{radn}\)
    \item \([1,1,0,0]\) to output for \(x_t^\mathrm{diff}\) based on \(x_1,\dots,x_{t-1}\) and \(x_t^\mathrm{radn},x_t^\mathrm{mint}\)
    \item \([1,1,1,0]\) to output for \(x_t^\mathrm{rain}\) based on \(x_1,\dots,x_{t-1}\) and \(x_t^\mathrm{radn},x_t^\mathrm{mint},x_t^\mathrm{diff}\).
\end{itemize}
Since the column size of 2 shrinks the length \(t_0\) by 1, we pad a single 0 after the convolution. We regard the masked convolution akin to data preprocessing, so we use neither bias term nor non-linear activation at this layer.

Next, each output of the masked convolution is processed by the dilated causal convolution. Our design principle is:
\begin{enumerate}
\item Choose \(T-t_0\), the length of future weather to generate.
\item Choose the acceptable range for \(t_0\), the target length of past observations based on which the future weather is generated.
\item Choose the length of the base filter and the number of convolutional layers in tandem with the acceptable range for \(t_0\).
\item Choose the number of channels based on the size of the training data and the available computational resources.
\end{enumerate}
Step 1 should reflect the purpose of weather generation. In our case, it is to improve accuracy and precision of crop yield prediction over a few years in the future by feeding the simulator with improved weather inputs. So, the target length is 365, 730 or 1095 days. Next, step 2 decides on how many past observations supply enough information for modelling the specified length of future weather. For example, El Niño and La Niña have significant impact on agricultural decision making and occur every two to seven years \cite{noaa_national_oceanic_and_atmospheric_administration_what_2024}. So, we may set \(t_0=365\times 6\) as the target value to ensure the acceptable range.

Step 3 typically requires some trial-and-error on the base filter length and the number of convolutional layers. As indicated in Section \ref{sec:conv}, the size of the receptive field exponentially grows as the number of layers grows. Let \(l\) be the length of the base filter and \(m\) be the number of convolutional layers. Then, the size of the receptive field is \(l^m\). When relating \(T-t_0\), \(t_0\) and \(l^m\), we may think as if there is only a single layer with a filter of size \(l^m\) that slides \(T-t_0-1\) times over the input of length \(t_0\) with the \(T-t_0-1\) padded 0s to compute the output of length \(T-t_0\). Thus, as already implied by the number padded 0s, the relation is \(l^m = T - 1\) and \(l^m - t_0\) number of padded 0s. With this constraint in mind, we decide on \(l\) and \(m\) to have \(t_0\) within the acceptable range.

Following the dilated convolutional layers, we use four \(1\times 1\) convolutional layers. Note that, the \(1\times 1\) convolutional layer is effectively the fully-connected layer but implemented convolutionally for the benefit of being fully convolutional networks \cite{long_fully_2015}. Regarding activation functions, we use \(\mathrm{tanh}\) after each of \(m\) dilated convolutional layers, \(\mathrm{ReLU}\) after each of \(1\times 1\) convolutional layers (except the last one), and the combination of identity and \(\mathrm{softplus}\) activation after the last layer (Section \ref{sec:loss_fn}).

Finally, as standard in convolutional neural network architectures, each convolutional layer outputs multiple channels. The common trade-off is applicable here: an increase in the non-linear approximation capacity of neural networks requires an increase in training data size and computing power. Note that, as we model time series, all the filters are in 1-D except the filters for the masked convolution, which are in 2-D for ease of PyTorch implementation.

\subsection{Construction of the training data}\label{sec:dataset}
\(t_0\) and \(T-t_0\) are the time-lengths of input \(x\) and target \(y\) respectively as a single training example. The training data contains many training examples of this kind. To construct the training data from the original time series dataset, we slice the original dataset into the time-length of \(T\) at many different points. For example, with \(T=2401\) and \(47847\) as the length of the portion of the original dataset used for training, we have 45446 distinct points to take a slice of length 2401, resulting in a training data that contains 45446 examples.

As a standard practice in deep learning, input \(x\) in the training set is standardised so that all the input variables have mean 0 and standard deviation of 1. (These sample means and standard deviations are also used to standardised input \(x\) in the test set.) Finally, given the use of gamma distributions for \texttt{radn}, \texttt{diff} and \texttt{rain}, we replace 0 (very common for \texttt{rain} and almost never for the others) with \(10^{-3}\) for numerical stability.

\section{Experiments}\label{sec:experiments}
The purpose of experiments are twofold:
\begin{enumerate}
\item Characterise the generated weather and compare it against the true weather in the test set.
\item Run crop model simulations using the generated weather and the one prepared by the conventional method, and compare their yield prediction errors from crop yield simulated using the true weather.
\end{enumerate}

For the conventional method, we chose one of the most representative approaches reviewed in Section \ref{sec:intro}. That is, a sequence of weather inputs fed into APSIM models was simply a sequence found in the past 30 years of historical observations, and the starting date (with the common month and day) of each sequence was randomly selected in one of 30 years. The other details specific to the scenarios are found in the next section.

By both methods, we drew 1000 random weather samples (i.e., 1000 sets of 4-dimensional vectors over the simulated days) to compute Monte Carlo estimates of quantities of interest. For reproducibility, random seeds were set equal to \texttt{1} in NumPy and PyTorch separately at the beginning of training and the beginning of sampling. All the code and APSIM simulation files (\texttt{.apsimx} files) are available on the \href{https://github.com/ysaikai/GenWeather}{website}.

\subsection{Scenarios}
We experimented two representative scenarios: single-year production using one of three crops (Scenario 1) and three-year production using rotations of the same three crops (Scenario 2). The original dataset was divided into two subsets: one for training and the other for testing (see details in the following sections). The simulated days were those contained in the test set, which were simulated separately in Scenario 1 and jointly in Scenario 2. Consequently, the lengths of future weather \((T-t_0)\) sampled by each method were 365 and 1095 in Scenario 1 and 2 respectively. Specifically for the conventional method, to sample a set of 365 vectors for the common starting month and day, it randomly picks a starting year from 1991--2020 for 2021, from 1992--2021 for 2022 and from 1993--2022 for 2023. The technical details of choosing the size of receptive field \((T-1)\) and the length of past observation \((t_0)\) are described in Section \ref{sec:architecture_specs}.

The crops used in both scenarios are wheat, barley and canola, whose yield was separated simulated in Scenario 1 and jointly simulated as part of a rotation in Scenario 2. The three-year rotations in Scenario 2 are 6 permutations of the crops:
\begin{itemize}
\item (Wheat, Barley, Canola)
\item (Wheat, Canola, Barley)
\item (Barley, Canola, Wheat)
\item (Barley, Wheat, Canola)
\item (Canola, Wheat, Barley)
\item (Canola, Barley, Wheat).
\end{itemize}

Note that we trained two separate neural networks for two scenarios. Although it is tempting to repurpose the network trained under Scenario 1 for Scenario 2 and may end up with reasonable values, we do not think this is a principled way to address Scenario 2. We should take samples from an estimate of the correct distribution which is defined in Equation \ref{eq:target}. If repurposing, samples of \(x_t\) for \(t\geq366+t_0\) would be taken from
\[
p(x_t|x_{t-t_0-364},\dots,x_{t-1})
\]
which is different from the correct distribution
\[
p(x_t|x_1,\dots,x_{t-1})
\]
because the size of the receptive field in Scenario 1 is only \(t_0+364\).

\subsection{Location and original dataset}
As emphasised in Section \ref{sec:intro}, training task-specific models is a key feature of our method. We used a single time series from a single location, Robe (South Australia), as its dataset contains daily observations of solar radiation (\texttt{radn}), minimum temperature (\texttt{mint}), maximum temperature (\texttt{maxt}) and rainfall (\texttt{rain}) over more than 100 years \cite{queensland_government_get_2024}. On 18 March 2024, we downloaded the dataset for 1 January 1890--17 March 2024, which contains 49019 daily records. As noted above, we divided it into two subsets: one containing the data from 1 January 1890 to 18 March 2021 for training and the other containing the data from 19 March 2021 to 17 March 2024 for testing.

\subsection{Crop model specifications}
To create crop models, we used APSIM Next Generation build 2023.11.7349.0. Given the chosen location (Robe, South Australia), we modified the default specifications of the following simulation files included in the downloaded APSIM software, \texttt{Wheat.apsimx}, \texttt{Barley.apsimx}, and \texttt{Canola.apsimx}. Specifically, we replaced the default soil profile with the one for Robe. We set the initial soil water at 20\% full. In terms of fertiliser management, 350kg/ha of nitrogen in urea was given at sowing of each crop in order to ensure non-limiting nitrogen supply \cite{peake_effect_2018} and focus on the effect of weather inputs to yield simulation. As implied throughout the paper, except weather inputs, all the crop model specifications are common between crops models by our method, the conventional method and the true weather. The complete specifications are found in \texttt{wheat.apsimx}, \texttt{barley.apsimx}, and \texttt{canola.apsimx} for Scenario 1 and \texttt{rotation.apsimx} for Scenario 2 available on the \href{https://github.com/ysaikai/GenWeather}{website}.

\subsection{Architecture specifications}\label{sec:architecture_specs}
Following the design principle (Section \ref{sec:architecture}), to ensure similar sizes of conditioning part based on which future weather is generated, we decided on
\begin{itemize}
    \item Scenario 1: \(m=4\) dilated convolutional layers with base filter of length \(l=7\),
    \item Scenario 2: \(m=5\) dilated convolutional layers with base filter of length \(l=5\).
\end{itemize}
Given the lengths of future weather \((T-t_0)\), 365 and 1095 in Scenario 1 and 2 respectively, the formula (Section \ref{sec:architecture}) gives us the following size of receptive field \((l^m)\) and the length of conditioning part \((t_0)\) in each scenario:
\begin{itemize}
    \item Scenario 1: \(l^m=T-1=2401\) and \(t_0=2037\)
    \item Scenario 2: \(l^m=T-1=3125\) and \(t_0=2031\).
\end{itemize}
Note similar \(t_0\) between two scenarios. Given these specifications, we processed the first subset of the data for training by the method described in Section \ref{sec:dataset} and constructed the training data that contained 45523 training examples in Scenario 1 and 44799 training examples in Scenario 2.

In Scenario 1, the number of output channels were \((8,16,32,64,64,32,16,8,2)\).
\begin{itemize}
\item (8): for the masked convolutional layer
\item (16,32,64,64): for the dilated convolutional layers
\item (32,16,8,2): for the 1×1 convolutional layers
\end{itemize}
The number \(2\) at the last layer is the number of parameters used to specify each of four distributions (Section \ref{sec:likelihood}). The other numbers were heuristically chosen by desired complexity of the network, learnability, computational requirement and empirical loss. In Scenario 2, the number of output channels were \((8,8,16,32,64,64,32,16,8,2)\), i.e., we simply inserted another dilated convolutional layer with 8 output channels. The resulting numbers of trainable parameters are 50682 in Scenario 1 and 37442 in Scenario 2. Note that, despite the smaller number of trainable parameters, the training in Scenario 2 took longer than in Scenario 1 due to larger \(T\).

\subsection{Other details}
We used Adam \cite{kingma_adam_2017} to carry out the neural network parameter optimisation. The learning rate was the default value (0.001) in PyTorch. As emphasised, our neural network is lightweight for the practical reasons (Section \ref{sec:intro}), and we trained it on a desktop computer with Nvidia GeForce RTX 3060 GPU with 12GB RAM, which is entry-level and significantly underpowered by today's deep learning standard. The versions of key software are Python 3.11.5, PyTorch 2.2.0, and NumPy 1.26.3.

\section{Results}\label{sec:results}
The neural network training took approximately 26.2 sec/epoch in Scenario 1 and 44.3 sec/epoch in Scenario 2. To serve the purpose of helping individual crop modellers (Section \ref{sec:intro}) and keep the training time around 24 hours, we spent 2000 epochs in both scenarios. The model used for testing was the one that achieved the lowest empirical loss in each scenario (loss=2.76 in Scenario 1 and loss=2.71 in Scenario 2). The trained PyTorch models (\texttt{scenario1.pt} and \texttt{scenario2.pt}) and the generated weather files (\texttt{.met} files) are available on the \href{https://github.com/ysaikai/GenWeather}{website}.

\subsection{Scenario 1}
1000 sets of 365 daily weather values from 19 March 2021, 2022 and 2023 were generated based on \(t_0=2037\) immediately preceding observations, i.e., from 21 August 2015 to 18 March 2021, from 20 August 2016 to 18 March 2022 and from 20 August 2017 to 18 March 2021 respectively. The same number of values over the same periods were also sampled using the conventional method. Using these weather values as inputs, wheat, barley and canola yield were separately simulated in each of 2021, 2022 and 2023.

\subsubsection{Generated weather}
Table \ref{table:weathersummary1} provides a summary of the generated weather. Details are found in the appendix. The entries in the tables are daily absolute differences, calculated by applying different smoothing periods for each year and each variable as follows.
\begin{enumerate}
\item For each of 1000 Monte Carlo samples, take the average separately of the generated values and the true values over the number of days in a given period. That is, average the values over 7 days for each week for ``Week'' period and average the values over the number of days in each month for ``Month'' period.
\item Take the absolute difference between each of the averaged generated values and each of the averaged true values.
\item Take the average of each difference over 1000 Monte Carlo samples.
\item Take the average of the weekly values over 52 weeks to get the entries for ``Week'' period and the monthly values over 12 months to get the entries for ``Month'' period.
\end{enumerate}
Table \ref{table:weather2021}, \ref{table:weather2021} and \ref{table:weather2021} in the appendix omit Step \#4 and contain the values for each week and each month.

\begin{table}[H]
\centering
\caption{Summary of daily absolute differences of the generated weather and the true weather in Scenario 1. In general, as the smoothing period becomes longer, the errors decrease. The specific way to calculate each entry is described in the main text above.}
\label{table:weathersummary1}
\vspace{7pt}
\begin{tabular}{llrrrr}
Year & Period & Radn & MinT & MaxT & Rain \\\hline
\multirow{3}{*}{2021} & Day & 4.93 & 2.97 & 3.07 & 2.63\\
 & Week & 2.56 & 1.76 & 1.89 & 1.73\\
 & Month & 1.87 & 1.25 & 1.14 & 1.03\\
\hline
\multirow{3}{*}{2022} & Day & 4.86 & 2.88 & 2.95 & 2.73\\
 & Week & 2.32 & 1.44 & 1.82 & 1.70\\
 & Month & 1.32 & 0.75 & 0.93 & 0.97\\
\hline
\multirow{3}{*}{2023} & Day & 4.65 & 2.99 & 2.83 & 2.41\\
 & Week & 2.34 & 1.79 & 1.68 & 1.54\\
 & Month & 1.51 & 1.21 & 1.03 & 1.00
\end{tabular}
\end{table}

\subsubsection{Crop yield}
Table \ref{table:APSIMerrors1} provides 18 performance metrics for our method and the conventional method. Each performance metrics is calculated as follows.
\begin{enumerate}
\item For each year and each crop, run 1000 simulations with weather inputs prepared by each method and record yield.
\item For each year and each crop, run a simulation with weather inputs of the true weather.
\item Take the absolute difference between each of 1000 yield and the true yield.
\item Calculate mean and standard deviation of the absolute differences.
\end{enumerate}

\begin{table}[H]
\centering
\caption{Mean and standard deviation of yield prediction errors (kg/ha) in Scenario 1. Our method outperformed the conventional method in every one of 18 cases.}
\label{table:APSIMerrors1}
\vspace{7pt}
\begin{tabular}{rlrrrrrr}
& &\multicolumn{2}{c}{Wheat}&\multicolumn{2}{c}{Barley}&\multicolumn{2}{c}{Canola} \\\hline
 Year & Method &  Mean &  STD &  Mean &  STD &  Mean &  STD \\\hline
\multirow{2}{*}{2021} & Generative & 525 & 436 & 783 & 559 & 841 & 655 \\
 & Conventional & 747 & 491 & 1,100 & 722 & 1,277 & 983 \\\hline
\multirow{2}{*}{2022} & Generative & 1,367 & 659 & 1,792 & 932 & 1,310 & 821 \\
 & Conventional & 1,421 & 782 & 2,175 & 1,318 & 1,866 & 1,135 \\\hline
\multirow{2}{*}{2023} & Generative & 820 & 648 & 1,523 & 1,030 & 2,038 & 987 \\
 & Conventional & 1,198 & 737 & 1,993 & 1,234 & 2,163 & 1,096
\end{tabular}
\end{table}

\subsection{Scenario 2}
1000 sets of 1095 daily weather values from 19 March 2021 were generated based on \(t_0=2031\) immediately preceding observations, i.e., from 27 August 2015 to 18 March 2021. The same number of values over the same period were also sampled using the conventional method. Using these weather values as inputs, 6 three-year rotations (i.e., 6 permutations of wheat, barley and canola) were simulated.

\subsubsection{Generated weather}
Table \ref{table:weathersummary2} provides a summary of the generated weather. Details are found in the appendix. The entries in the table are daily absolute differences, calculated by applying different smoothing periods for each variable as follows.
\begin{enumerate}
\item For each of 1000 Monte Carlo samples, take the average separately of the generated values and the true values over the number of days in a given period. That is, average the values over 7 days for each week for ``Week'' period and average the values over the number of days in each month for ``Month'' period. Note that each of 52 weeks and 12 months occurs 3 times over 1095 days.
\item Take the absolute difference between each of the averaged generated values and each of the averaged true values.
\item Take the average of each difference over 1000 Monte Carlo samples.
\item Take the average of the weekly values over 52 weeks to get the entries for ``Week'' period and the monthly values over 12 months to get the entries for ``Month'' period.
\end{enumerate}
Table \ref{table:weather2021-2023} in the appendix omits Step \#4 and contain the values for each week and each month.

\begin{table}[H]
\centering
\caption{Summary of daily absolute differences of the generated weather and the true weather over 2021--2023 in Scenario 2. In general, as the smoothing period becomes longer, the errors decrease. The specific way to calculate each entry is described in the main text above.}
\label{table:weathersummary2}
\vspace{7pt}
\begin{tabular}{lrrrr}
Period & Radn & MinT & MaxT & Rain \\\hline
Day & 4.85 & 2.95 & 3.09 & 2.68\\
Week & 2.05 & 1.14 & 1.40 & 1.13\\
Month & 1.91 & 0.88 & 1.09 & 0.67
\end{tabular}
\end{table}

\subsubsection{Crop yield}
Table \ref{table:APSIMerrors2} provides 36 performance metrics for our method and the conventional method. Each performance metrics is calculated as follows.
\begin{enumerate}
\item For crop rotation, run 1000 simulations with weather inputs prepared by our method and the conventional method and record yield.
\item For each crop rotation, run a simulation with weather inputs of the true weather.
\item For each crop in each rotation, take the absolute difference between each of 1000 yield and the true yield.
\item Calculate mean and standard deviation of the absolute differences.
\end{enumerate}

\begin{table}[H]
\centering
\caption{Mean and standard deviation of yield prediction errors (kg/ha) in Scenario 2. The shaded values imply where our method had inferior performance than the conventional method. Overall, our method outperformed the conventional method in 29 out of 36 cases.}
\label{table:APSIMerrors2}
\vspace{7pt}
\begin{tabular}{llrrrrrr}
 & &\multicolumn{2}{c}{Wheat}&\multicolumn{2}{c}{Barley}&\multicolumn{2}{c}{Canola} \\\hline
Rotation &       Method & Mean & STD & Mean & STD & Mean & STD \\\hline
\multirow{2}{*}{WBC} & Generative & 893 & \colorbox{black!10}{604} & 736 & 660 & 672 & 491 \\
 & Conventional & 896 & \colorbox{black!10}{472} & 1720 & 1245 & 746 & 507 \\\hline
\multirow{2}{*}{WCB} & Generative & 893 & \colorbox{black!10}{604} & 1127 & \colorbox{black!10}{708} & 969 & 704 \\
 & Conventional & 914 & \colorbox{black!10}{471} & 1290 & \colorbox{black!10}{703} & 1272 & 900 \\\hline
\multirow{2}{*}{BWC} & Generative & 911 & 641 & 887 & 624 & 680 & 492 \\
 & Conventional & 1781 & 1020 & 1074 & 864 & 827 & 562 \\\hline
\multirow{2}{*}{BCW} & Generative & 948 & 666 & 887 & 624 & 878 & 651 \\
 & Conventional & 1162 & 686 & 1051 & 844 & 1112 & 712 \\\hline
\multirow{2}{*}{CWB} & Generative & 929 & 608 & 1142 & 711 & \colorbox{black!10}{1087} & \colorbox{black!10}{502} \\
 & Conventional & 1800 & 996 & 1234 & 711 & \colorbox{black!10}{618} & \colorbox{black!10}{480} \\\hline
\multirow{2}{*}{CBW} & Generative & 971 & 689 & 722 & 640 & \colorbox{black!10}{1087} & \colorbox{black!10}{502} \\
 & Conventional & 1174 & 713 & 1734 & 1238 & \colorbox{black!10}{602} & \colorbox{black!10}{495}
\end{tabular}
\end{table}

\section{Discussion}\label{sec:discussion}
Although the ultimate goal is to improve crop model simulations using generative weather inputs, there are a few comments to make on the generated weather itself as intermediate results. First, the weather seasonality within each year is clearly captured by the learned models. For example, in Table \ref{table:weather2021}, \ref{table:weather2022}, \ref{table:weather2023} and \ref{table:weather2021-2023}, the magnitude of errors in minimum temperature and maximum temperature are consistent throughout each year, implying temperature is correctly high in the summer months (December, January and February in the Southern Hemisphere) and correctly low in the winter months (June, July and August in the Southern Hemisphere). Also indicated in these tables is that the errors are smaller in month than in week due to the greater smoothing (i.e., averaging over 30 days compared to averaging over 7 days before taking the difference.)

As mentioned in Section \ref{sec:likelihood}, since rain is a sparse and sporadic event at this location (no rain on more than 50\% of the recorded days and occasional heavy rain), we expected that errors in rainfall might be unacceptably high. However, as found in Table \ref{table:weathersummary1} and \ref{table:weathersummary2}, this is not the case; the daily error was at most 2.73, which we consider acceptable for inputs to crop models. We suspect that, albeit many, incorrectly predicted rainy days did not contribute much to the overall error because those positive rainfalls are small. Since exactly 0 mm rainfall is a measure-zero event for continuous distributions, generation of 0s happens only because of numerical rounding. If some applications require greater control of generating 0s, it may be worthwhile to discretise continuous rainfalls with the softmax distribution (e.g., PixelCNN \cite{van_den_oord_pixel_2016} and WaveNet \cite{van_den_oord_wavenet_2016}) or to use a mix of discrete and continuous distributions (i.e., Bernoulli distribution of no rain and a continuous distribution conditional on rain). The latter is simpler, requiring to estimate only one more parameter for Bernoulli distribution per time step, so may be more suitable for lightweight neural networks.

Since accuracy and precision are of major concern when crop models are used for yield prediction, we have measured the performance of our method and the conventional method in terms of mean and standard deviation of prediction errors. Thus, interpretation of results is simple---the smaller, the better---because smaller mean implies higher accuracy, and smaller standard deviation implies higher precision. In Scenario 1, the results are extreme because our method outperformed the conventional in every one of 18 metrics. In Scenario 2, the results are more realistic as our method outperformed the conventional in 29 out of 36 metrics. A conjecture why the performance in Scenario 2 decreases is the following. Although the length of the generated weather in Scenario 2 is three times as long as the length of the one in Scenario 1, the neural network in Scenario 2 is actually simpler in terms of the number of trainable parameters (Section \ref{sec:architecture_specs}). We made this compromise because of the target training hours (24 hours using desktops), which is justified in Section \ref{sec:intro}. The less complex network may be a cause of the lesser performance, but we need to investigate the matter in many more different scenarios.

Our method is significantly more flexible than the conventional method for weather input preparation. As explained in Section \ref{sec:architecture}, the method allows modellers to choose desired lengths of past observations and future weather data depending on the specificity of their problems and available data \& computational resources. Notice that the borderline between the past observations and the future weather data need not be a point in time between two cropping seasons; it can be in the middle of the season. (The borderlines used in the experiments, 18 March 2021, 2022 and 2023, are merely because we downloaded the dataset that contained the data up to 17 March 2024 and used all of them in both scenarios.) For example, the method can be used to generate weather from a day in the season to the end of the season based on the observations up until the day before. An implication is that crop modellers can update the yield forecast in a season of unusual weather or make in-season management choices to adapt to an unusual season. This type of flexible modelling is difficult or at least unsound if simply using historical records without explicit modelling as the conventional method does. For example, Yield Prophet, an APSIM-based decision support tool, makes an arbitrary switch from the realised weather sequence to one of the historical sequences over the past 100 years in the middle of the season \cite{hochman_re-inventing_2009}. We consider the practice scientifically unsound because it likely breaks the continuity of the atmospheric process, i.e., the temporal dependency.

To simplify the exposition, we used a single time series for a single location as the original dataset. Modellers may apply the procedure explained in Section \ref{sec:dataset} to multiple series and create training examples from different series. While this is a simple way to increase the size of the training data as weather data is readily available for many locations, modellers should be careful about using weather data for locations that have distinct climates---similar care taken in transfer learning \cite{pan_survey_2010}. As emphasised in Section \ref{sec:intro}, our method takes advantage of the task-specific nature of machine learning models for weather generation. For example, if the task is to generate weather for location A, use of data for location B to increase the size of the training data for this task is not necessarily advantageous because location B may have a very different climate and use of its data may mislead the model training.

Design of neural network architectures is where modellers make trade-offs between pros and cons of using convolutional networks for sequence modelling. For the benefits emphasised in Section \ref{sec:intro} (i.e., simpler and cleaner architectures than recurrent networks), modellers must accept lower flexibility---size of data to generate is part of the network architecture. This is not the case in recurrent networks, where we can sequentially sample data of arbitrary size using a trained model. As explained in Section \ref{sec:architecture}, the desired size of outputs \((T-t_0)\) is jointly determined with the size of the receptive field \((l^m=T-1)\), which is a major architectural component of convolutional networks. For the purpose of generating weather inputs to crop models, the constraint is hardly an issue because crop modellers typically have specific numbers of days to run crop simulations for. This is a reason why we chose the fully convolutional network for generating sequences of weather data.

As to the specific architectures adopted for the experiments (Section \ref{sec:architecture_specs}), we followed the design principle explained in Section \ref{sec:architecture}. First, we chose two desired numbers of days \((T-t_0)\) to run crop simulations for as part of the scenario specifications. Then, we heuristically chose the acceptable range for the length of conditioning part around \(t_0\approx 2000\). We think that a reasonable \(t_0\) can be common in different scenarios because, when iteratively generating a sequence of data, values at the first time step \(x_{t_0+1}\) is crucial as they influence all the subsequent values, and \(t_0\) reflects how much information in the past is used to generate \(x_{t_0+1}\) regardless of the future length \((T-t_0)\). Needless to say, \(t_0\approx 2000\) is by no means a magic number and should be set according to characteristics of particular applications, network architectures and available computational resources, because the larger \(t_0\), the more computation required. Finally, to achieve the target \(t_0\approx 2000\) in both scenarios, we chose the numbers of dilated convolutional layers and the sizes of base filter (i.e., \(m=4\) and \(l=7\) in Scenario 1 and \(m=5\) and \(l=5\) in Scenario 2).

There is another flexibility in our method that crop modellers may take advantage of. Recall that, within each time step \(t\), the probability distribution to estimate is broken into a sequence of four conditional distributions (Equation \ref{eq:likelihood}), resulting in the additively separable loss function among 4 weather variables (Equation \ref{eq:loss_fn}). For example, if some weather variable has a critical threshold above which irreversible damage occurs to the crop (e.g., maximum temperature above which the crop loses the economic value), modellers may want to generate samples of that variable with greater accuracy and precision than of the other variables. In this case, they can simply multiply the corresponding term in the loss function by a number greater than 1. This way, the neural network training will be steered towards reducing more of the loss contributed by that variable.



\section{Conclusion}\label{sec:conclusion}
To address the lack of advanced methods for preparing weather inputs to crop model simulations, we have proposed the new method that enables crop modellers to train unique neural networks and generate weather inputs to their crop models. As demonstrated, quality of weather inputs has a significant impact on crop model simulations, which are often used for crop yield prediction. Under the circumstances of the rising global population and the deteriorating natural environment, accurate and precise yield prediction is more crucial than ever for effective decision making. We hope the method helps crop modellers improve yield prediction, thereby more effectively addressing this existential threat to the world.

\section*{Acknowledgements}
The author is grateful to Matthew Knowling who provided helpful comments.

\section*{Appendix}
The entries in the tables are daily absolute differences, calculated by applying different smoothing periods for each year and each variable as follows.
\begin{enumerate}
\item For each of 1000 Monte Carlo samples, take the average separately of the generated values and the true values over the number of days in a period. That is, average the values over 7 days for each week for ``Week'' and average the values over the number of days in each month for ``Month''.
\item Take the absolute difference between each of the averaged generated values and each of the averaged true values.
\item Take the average of each difference over 1000 Monte Carlo samples.
\end{enumerate}

\newpage
\begin{table}[H]
\centering
\caption{Estimated daily absolute errors in 2021.}
\label{table:weather2021}
\begin{tabular}{lrrrr}
Period & Radn & MinT & MaxT & Rain \\\hline
January & 1.86 & 2.79 & 3.11 & 0.32 \\\hline
February & 2.05 & 0.95 & 1.12 & 1.2 \\\hline
March & 3.21 & 1.62 & 1.83 & 1.07 \\\hline
April & 0.61 & 0.83 & 0.72 & 0.76 \\\hline
May & 1.36 & 1.47 & 0.82 & 1.2 \\\hline
June & 0.38 & 1.65 & 0.5 & 2.24 \\\hline
July & 0.28 & 0.87 & 0.68 & 1.56 \\\hline
August & 1.61 & 1.72 & 0.73 & 1.11 \\\hline
September & 1.9 & 0.69 & 1.02 & 1.16 \\\hline
October & 2.1 & 0.77 & 0.78 & 0.68 \\\hline
November & 2.22 & 1.06 & 1.27 & 0.48 \\\hline
December & 4.88 & 0.6 & 1.07 & 0.54
\end{tabular}
\end{table}

\begin{minipage}{0.5\linewidth}
\centering
\begin{tabular}{lrrrr}
Period & Radn & MinT & MaxT & Rain \\\hline
Week 01 & 6.1 & 1.76 & 1.96 & 0.38 \\\hline
Week 02 & 2.45 & 1.86 & 4.76 & 0.6 \\\hline
Week 03 & 2.8 & 1.7 & 2.81 & 0.41 \\\hline
Week 04 & 3.53 & 7.37 & 8.11 & 0.58 \\\hline
Week 05 & 2.05 & 1.43 & 2.34 & 0.68 \\\hline
Week 06 & 3.76 & 1.78 & 3.67 & 1.29 \\\hline
Week 07 & 1.84 & 1.51 & 1.86 & 0.93 \\\hline
Week 08 & 4.7 & 1.6 & 1.74 & 1.46 \\\hline
Week 09 & 2.22 & 3.25 & 4.14 & 1.59 \\\hline
Week 10 & 3.67 & 1.23 & 1.96 & 0.95 \\\hline
Week 11 & 3.99 & 1.88 & 3.09 & 1.07 \\\hline
Week 12 & 2.02 & 2.19 & 2.86 & 1.91 \\\hline
Week 13 & 4.61 & 1.64 & 2.23 & 1.57 \\\hline
Week 14 & 2.1 & 1.15 & 2.38 & 2.14 \\\hline
Week 15 & 1.69 & 1.45 & 1.07 & 1.41 \\\hline
Week 16 & 2.2 & 1.51 & 1.98 & 2.69 \\\hline
Week 17 & 1.54 & 1.29 & 1.08 & 1.7 \\\hline
Week 18 & 2.2 & 1.25 & 1.79 & 4.11 \\\hline
Week 19 & 1.16 & 1.69 & 0.87 & 1.33 \\\hline
Week 20 & 0.89 & 2.98 & 0.96 & 1.6 \\\hline
Week 21 & 1.86 & 3.27 & 2.44 & 1.44 \\\hline
Week 22 & 1.27 & 1.95 & 0.96 & 2.07 \\\hline
Week 23 & 0.71 & 2.55 & 0.95 & 5.62 \\\hline
Week 24 & 1.06 & 1.95 & 0.89 & 3.75 \\\hline
Week 25 & 0.62 & 1.56 & 0.98 & 2.45 \\\hline
Week 26 & 0.75 & 1.9 & 0.74 & 2.06
\end{tabular}
\end{minipage}
\begin{minipage}{0.5\linewidth}
\centering
\begin{tabular}{lrrrr}
Period & Radn & MinT & MaxT & Rain \\\hline
Week 27 & 0.57 & 1.4 & 1.13 & 2.51 \\\hline
Week 28 & 0.7 & 1.28 & 0.69 & 2.47 \\\hline
Week 29 & 0.86 & 1.09 & 1.05 & 5.99 \\\hline
Week 30 & 1.13 & 1.51 & 0.68 & 2.26 \\\hline
Week 31 & 1.05 & 1.8 & 0.61 & 2.08 \\\hline
Week 32 & 1.13 & 2.03 & 1.07 & 1.73 \\\hline
Week 33 & 2.24 & 2.53 & 1.39 & 2.79 \\\hline
Week 34 & 2.34 & 1.92 & 1.17 & 1.62 \\\hline
Week 35 & 1.3 & 1.1 & 1.33 & 1.43 \\\hline
Week 36 & 2.46 & 1.36 & 0.92 & 1.6 \\\hline
Week 37 & 2.05 & 1.3 & 1.13 & 1.56 \\\hline
Week 38 & 3.48 & 0.89 & 0.95 & 2.01 \\\hline
Week 39 & 1.24 & 1.47 & 2.15 & 1.37 \\\hline
Week 40 & 2.91 & 1.45 & 1.23 & 2.69 \\\hline
Week 41 & 2.43 & 2.09 & 1.89 & 0.97 \\\hline
Week 42 & 2.23 & 1.54 & 1.53 & 0.73 \\\hline
Week 43 & 4.36 & 0.93 & 2.07 & 1.59 \\\hline
Week 44 & 5.99 & 1.12 & 1.74 & 1.71 \\\hline
Week 45 & 2.38 & 1.83 & 2.24 & 0.93 \\\hline
Week 46 & 2.87 & 2.46 & 3.8 & 1.14 \\\hline
Week 47 & 2.66 & 1.1 & 1.85 & 0.9 \\\hline
Week 48 & 4.72 & 1.02 & 1.66 & 0.76 \\\hline
Week 49 & 4.96 & 1.27 & 2.23 & 1.25 \\\hline
Week 50 & 5.18 & 1.26 & 1.91 & 0.74 \\\hline
Week 51 & 5.19 & 1.15 & 1.74 & 1.03 \\\hline
Week 52 & 4.71 & 1.09 & 1.64 & 0.36
\end{tabular}
\end{minipage}

\newpage
\begin{table}[H]
\centering
\caption{Estimated daily absolute errors in 2022.}
\label{table:weather2022}
\begin{tabular}{lrrrr}
Period & Radn & MinT & MaxT & Rain \\\hline
January & 3.75 & 0.55 & 1.31 & 0.44 \\\hline
February & 0.96 & 0.63 & 1.03 & 0.71 \\\hline
March & 0.66 & 0.53 & 0.73 & 1.08 \\\hline
April & 0.57 & 0.67 & 1.79 & 1.73 \\\hline
May & 1.41 & 0.91 & 0.72 & 0.87 \\\hline
June & 0.42 & 1.24 & 0.52 & 1.08 \\\hline
July & 0.49 & 0.69 & 0.41 & 1.25 \\\hline
August & 0.45 & 0.7 & 0.43 & 1.26 \\\hline
September & 1.22 & 0.6 & 0.83 & 0.76 \\\hline
October & 2.09 & 1.06 & 1.24 & 0.75 \\\hline
November & 1.13 & 0.7 & 0.84 & 1.26 \\\hline
December & 2.77 & 0.76 & 1.36 & 0.48
\end{tabular}
\end{table}

\begin{minipage}{0.5\linewidth}
\centering
\begin{tabular}{lrrrr}
Period & Radn & MinT & MaxT & Rain \\\hline
Week 01 & 7.6 & 1.18 & 4.45 & 0.63 \\\hline
Week 02 & 6.14 & 1.97 & 4.34 & 0.46 \\\hline
Week 03 & 2.23 & 1.33 & 2.44 & 1.29 \\\hline
Week 04 & 2.87 & 1.07 & 1.73 & 0.73 \\\hline
Week 05 & 1.98 & 1.73 & 3.44 & 3.39 \\\hline
Week 06 & 1.95 & 1.13 & 1.81 & 0.36 \\\hline
Week 07 & 2.17 & 1.1 & 1.87 & 0.47 \\\hline
Week 08 & 2.12 & 1.69 & 2.82 & 1.42 \\\hline
Week 09 & 1.52 & 1.19 & 1.95 & 1.51 \\\hline
Week 10 & 2.21 & 1.15 & 1.73 & 1.37 \\\hline
Week 11 & 1.59 & 1.06 & 2 & 1.34 \\\hline
Week 12 & 1.91 & 1.29 & 1.15 & 1.57 \\\hline
Week 13 & 2.84 & 0.97 & 1.37 & 1.62 \\\hline
Week 14 & 1.26 & 1.45 & 1.31 & 1.97 \\\hline
Week 15 & 1.32 & 1.12 & 1.75 & 2.13 \\\hline
Week 16 & 1.03 & 1.52 & 1 & 1.68 \\\hline
Week 17 & 1.18 & 2.04 & 4.53 & 2.77 \\\hline
Week 18 & 2.7 & 1.29 & 0.81 & 4.36 \\\hline
Week 19 & 1.36 & 1.45 & 1.36 & 1.12 \\\hline
Week 20 & 1.61 & 2.53 & 0.92 & 1.32 \\\hline
Week 21 & 1.96 & 1.35 & 1.63 & 1.84 \\\hline
Week 22 & 0.86 & 1.52 & 1.18 & 1.79 \\\hline
Week 23 & 0.55 & 2.04 & 1.02 & 3.94 \\\hline
Week 24 & 1.06 & 1.47 & 0.75 & 1.85 \\\hline
Week 25 & 0.47 & 2.06 & 0.75 & 1.97 \\\hline
Week 26 & 0.72 & 1.42 & 0.66 & 1.76
\end{tabular}
\end{minipage}
\begin{minipage}{0.5\linewidth}
\centering
\begin{tabular}{lrrrr}
Period & Radn & MinT & MaxT & Rain \\\hline
Week 27 & 0.84 & 1.14 & 0.59 & 2.36 \\\hline
Week 28 & 0.51 & 1.31 & 0.9 & 1.92 \\\hline
Week 29 & 0.94 & 1.48 & 0.6 & 1.63 \\\hline
Week 30 & 0.9 & 1.17 & 0.77 & 1.6 \\\hline
Week 31 & 1.4 & 0.9 & 0.75 & 1.95 \\\hline
Week 32 & 0.65 & 1.14 & 0.71 & 2.31 \\\hline
Week 33 & 1.06 & 1.16 & 0.8 & 1.03 \\\hline
Week 34 & 1.67 & 1.51 & 0.86 & 3.56 \\\hline
Week 35 & 1.11 & 1.09 & 0.95 & 1.51 \\\hline
Week 36 & 1.32 & 1.18 & 1.69 & 1.57 \\\hline
Week 37 & 1.51 & 1.02 & 1.01 & 2.05 \\\hline
Week 38 & 1.93 & 1.22 & 1.38 & 1.44 \\\hline
Week 39 & 2.73 & 1.29 & 1.23 & 1.36 \\\hline
Week 40 & 1.78 & 2.02 & 2.87 & 1.68 \\\hline
Week 41 & 3.13 & 1.08 & 1.6 & 1.33 \\\hline
Week 42 & 8.13 & 1.03 & 2.64 & 1.36 \\\hline
Week 43 & 3.24 & 2.49 & 1.61 & 1.73 \\\hline
Week 44 & 2.18 & 1.55 & 2.08 & 3.01 \\\hline
Week 45 & 4.38 & 2.87 & 4.62 & 0.84 \\\hline
Week 46 & 3.71 & 1.77 & 2.84 & 2.71 \\\hline
Week 47 & 3.21 & 1.8 & 1.68 & 2 \\\hline
Week 48 & 2.27 & 0.99 & 1.34 & 0.97 \\\hline
Week 49 & 4.04 & 1.52 & 1.39 & 1.33 \\\hline
Week 50 & 3.94 & 2.13 & 2.88 & 1.32 \\\hline
Week 51 & 2.37 & 0.92 & 3.58 & 0.69 \\\hline
Week 52 & 8.65 & 1.12 & 4.48 & 0.67
\end{tabular}
\end{minipage}

\newpage
\begin{table}[H]
\centering
\caption{Estimated daily absolute errors in 2023.}
\label{table:weather2023}
\begin{tabular}{lrrrr}
Period & Radn & MinT & MaxT & Rain \\\hline
January & 1.33 & 0.68 & 1.01 & 0.34 \\\hline
February & 1.83 & 1.19 & 1.02 & 0.74 \\\hline
March & 1.58 & 0.72 & 1.52 & 1.07 \\\hline
April & 0.72 & 1.2 & 0.82 & 1.2 \\\hline
May & 1.16 & 1.12 & 0.62 & 0.69 \\\hline
June & 0.5 & 2.51 & 0.56 & 2.63 \\\hline
July & 0.91 & 2.6 & 0.6 & 1.1 \\\hline
August & 1.6 & 1.3 & 0.8 & 1.88 \\\hline
September & 2.82 & 1.17 & 2.17 & 0.73 \\\hline
October & 1.64 & 0.67 & 0.93 & 0.59 \\\hline
November & 2.58 & 0.77 & 1.32 & 0.52 \\\hline
December & 1.43 & 0.61 & 0.94 & 0.56
\end{tabular}
\end{table}

\begin{minipage}{0.5\linewidth}
\centering
\begin{tabular}{lrrrr}
Period & Radn & MinT & MaxT & Rain \\\hline
Week 01 & 2.16 & 1.61 & 2.29 & 0.86 \\\hline
Week 02 & 1.89 & 1.33 & 1.75 & 0.47 \\\hline
Week 03 & 1.91 & 1.17 & 1.96 & 0.21 \\\hline
Week 04 & 1.81 & 1.23 & 1.96 & 0.47 \\\hline
Week 05 & 4.06 & 1.42 & 1.75 & 0.47 \\\hline
Week 06 & 3.64 & 1.47 & 2.21 & 0.85 \\\hline
Week 07 & 4.69 & 2.52 & 1.84 & 1.19 \\\hline
Week 08 & 3.78 & 1.4 & 1.88 & 0.37 \\\hline
Week 09 & 3.05 & 1.47 & 1.79 & 1.11 \\\hline
Week 10 & 4.02 & 1.05 & 3.34 & 0.3 \\\hline
Week 11 & 1.81 & 2.81 & 3.88 & 1.05 \\\hline
Week 12 & 2.46 & 1.01 & 1.5 & 0.98 \\\hline
Week 13 & 1.21 & 1.14 & 1.27 & 2.73 \\\hline
Week 14 & 2.38 & 1.3 & 1.42 & 0.81 \\\hline
Week 15 & 2.19 & 1.17 & 1.29 & 2.12 \\\hline
Week 16 & 1.82 & 2.32 & 1.41 & 1.56 \\\hline
Week 17 & 1.32 & 1.95 & 2 & 2.86 \\\hline
Week 18 & 1.98 & 2.46 & 1.3 & 1.63 \\\hline
Week 19 & 2.32 & 1.34 & 1.26 & 1.19 \\\hline
Week 20 & 3.01 & 1.31 & 0.82 & 2.12 \\\hline
Week 21 & 1.31 & 1.52 & 0.69 & 1.61 \\\hline
Week 22 & 0.82 & 2.38 & 0.83 & 2.38 \\\hline
Week 23 & 1.22 & 3.3 & 2.38 & 2.12 \\\hline
Week 24 & 0.67 & 3.78 & 1.09 & 3.85 \\\hline
Week 25 & 0.68 & 1.7 & 1.8 & 4.43 \\\hline
Week 26 & 1.11 & 2.91 & 0.75 & 3.27
\end{tabular}
\end{minipage}
\begin{minipage}{0.5\linewidth}
\centering
\begin{tabular}{lrrrr}
Period & Radn & MinT & MaxT & Rain \\\hline
Week 27 & 0.77 & 2.41 & 0.69 & 2.1 \\\hline
Week 28 & 1.05 & 4.53 & 1.52 & 1.5 \\\hline
Week 29 & 1.85 & 1.6 & 0.61 & 2.07 \\\hline
Week 30 & 1.35 & 1.82 & 0.78 & 1.79 \\\hline
Week 31 & 2.34 & 3.61 & 1.03 & 2.36 \\\hline
Week 32 & 1.13 & 1.4 & 0.77 & 1.86 \\\hline
Week 33 & 1.26 & 1.14 & 0.81 & 2.71 \\\hline
Week 34 & 2.24 & 2.51 & 1.27 & 1.47 \\\hline
Week 35 & 1.11 & 2.04 & 1.49 & 2.2 \\\hline
Week 36 & 2.23 & 2.55 & 1.82 & 2.52 \\\hline
Week 37 & 4.79 & 2.22 & 3.63 & 1.87 \\\hline
Week 38 & 2.24 & 1.3 & 1.79 & 1.07 \\\hline
Week 39 & 3.31 & 1.57 & 2.16 & 1.28 \\\hline
Week 40 & 1.36 & 1.26 & 1.44 & 1.26 \\\hline
Week 41 & 3.06 & 1.23 & 1.68 & 1.15 \\\hline
Week 42 & 2.1 & 1.34 & 2.47 & 0.99 \\\hline
Week 43 & 2.25 & 1.41 & 1.39 & 0.83 \\\hline
Week 44 & 3.65 & 1.44 & 2.34 & 0.99 \\\hline
Week 45 & 4.71 & 1.51 & 1.72 & 0.82 \\\hline
Week 46 & 2.44 & 1.25 & 2.3 & 1.42 \\\hline
Week 47 & 2.29 & 1.6 & 1.73 & 1.71 \\\hline
Week 48 & 4.1 & 1.13 & 2.29 & 0.75 \\\hline
Week 49 & 2.33 & 1.3 & 1.93 & 0.82 \\\hline
Week 50 & 2.6 & 1.09 & 1.9 & 1.81 \\\hline
Week 51 & 5.08 & 1.56 & 1.74 & 1.19 \\\hline
Week 52 & 2.6 & 1.09 & 1.68 & 0.52
\end{tabular}
\end{minipage}

\newpage
The entries in the table are daily absolute differences, calculated by applying different smoothing periods for each variable as follows.
\begin{enumerate}
\item For each of 1000 Monte Carlo samples, take the average separately of the generated values and the true values over the number of days in a period. That is, average the values over 7 days for each week for ``Week'' and average the values over the number of days in each month for ``Month''. Note that each of 52 weeks and 12 months occurs 3 times over 1095 days.
\item Take the absolute difference between each of the averaged generated values and each of the averaged true values.
\item Take the average of each difference over 1000 Monte Carlo samples.
\end{enumerate}

\newpage
\begin{table}[H]
\centering
\caption{Estimated daily absolute errors over 2021--2023.}
\label{table:weather2021-2023}
\begin{tabular}{lrrrr}
Period & Radn & MinT & MaxT & Rain \\\hline
January & 3 & 0.95 & 1.91 & 0.66 \\\hline
February & 2.5 & 0.58 & 0.65 & 0.66 \\\hline
March & 1.93 & 0.89 & 1.5 & 0.57 \\\hline
April & 0.83 & 1.21 & 1.64 & 0.51 \\\hline
May & 1.31 & 0.99 & 0.73 & 0.67 \\\hline
June & 0.89 & 1.26 & 0.6 & 1.71 \\\hline
July & 0.81 & 1.03 & 0.6 & 0.73 \\\hline
August & 0.96 & 1.47 & 1.34 & 0.65 \\\hline
September & 1.42 & 0.8 & 1.53 & 0.45 \\\hline
October & 1.64 & 0.54 & 1.38 & 0.49 \\\hline
November & 3.59 & 0.35 & 0.67 & 0.48 \\\hline
December & 3.99 & 0.51 & 0.54 & 0.41
\end{tabular}
\end{table}

\begin{minipage}{0.5\linewidth}
\centering
\begin{tabular}{lrrrr}
Period & Radn & MinT & MaxT & Rain \\\hline
Week 01 & 1.49 & 0.94 & 1.54 & 0.59 \\\hline
Week 02 & 4.42 & 1.27 & 3.47 & 0.71 \\\hline
Week 03 & 5.41 & 0.64 & 1.44 & 1.02 \\\hline
Week 04 & 1.65 & 2.28 & 2.61 & 0.84 \\\hline
Week 05 & 2.82 & 0.63 & 1.47 & 1 \\\hline
Week 06 & 3.25 & 0.79 & 1.08 & 0.93 \\\hline
Week 07 & 3.55 & 1.12 & 1.8 & 1.04 \\\hline
Week 08 & 2.83 & 0.71 & 1.7 & 0.84 \\\hline
Week 09 & 1.43 & 1.14 & 1.5 & 0.82 \\\hline
Week 10 & 2.86 & 0.76 & 1.27 & 0.68 \\\hline
Week 11 & 1.8 & 1.86 & 3.13 & 0.75 \\\hline
Week 12 & 1.49 & 0.88 & 1.64 & 0.83 \\\hline
Week 13 & 1.56 & 1.14 & 1.1 & 0.66 \\\hline
Week 14 & 1.01 & 0.72 & 1.6 & 0.96 \\\hline
Week 15 & 0.94 & 1.02 & 1.23 & 0.93 \\\hline
Week 16 & 1 & 2.78 & 1.55 & 1 \\\hline
Week 17 & 1.91 & 1.58 & 2.75 & 0.96 \\\hline
Week 18 & 0.76 & 1.15 & 0.65 & 1.5 \\\hline
Week 19 & 1.82 & 0.71 & 1.02 & 2.24 \\\hline
Week 20 & 1.39 & 1.85 & 0.88 & 1.33 \\\hline
Week 21 & 1.54 & 1.27 & 1.63 & 1.11 \\\hline
Week 22 & 1.08 & 0.76 & 0.46 & 1.1 \\\hline
Week 23 & 0.69 & 1.97 & 1.04 & 2.98 \\\hline
Week 24 & 1.37 & 1.57 & 1.3 & 1.79 \\\hline
Week 25 & 0.53 & 1.32 & 0.42 & 2.35 \\\hline
Week 26 & 1.52 & 1.23 & 0.47 & 1.85
\end{tabular}
\end{minipage}
\begin{minipage}{0.5\linewidth}
\centering
\begin{tabular}{lrrrr}
Period & Radn & MinT & MaxT & Rain \\\hline
Week 27 & 0.9 & 0.77 & 0.37 & 1.13 \\\hline
Week 28 & 0.8 & 1.27 & 0.87 & 1.97 \\\hline
Week 29 & 1.12 & 0.78 & 0.5 & 1.47 \\\hline
Week 30 & 0.82 & 1.42 & 1.11 & 1.15 \\\hline
Week 31 & 1.32 & 2.07 & 1.27 & 1.02 \\\hline
Week 32 & 0.53 & 0.67 & 0.87 & 1.04 \\\hline
Week 33 & 0.8 & 1.61 & 1.23 & 1.49 \\\hline
Week 34 & 1.39 & 1.85 & 1.27 & 1 \\\hline
Week 35 & 1.03 & 1.82 & 2.48 & 1.94 \\\hline
Week 36 & 0.8 & 1.21 & 1.17 & 0.91 \\\hline
Week 37 & 2.01 & 0.88 & 1.38 & 0.87 \\\hline
Week 38 & 0.9 & 1.08 & 1.25 & 0.76 \\\hline
Week 39 & 2.37 & 0.68 & 2.1 & 0.92 \\\hline
Week 40 & 0.85 & 1.51 & 1.74 & 0.91 \\\hline
Week 41 & 2.18 & 0.75 & 1.3 & 1.2 \\\hline
Week 42 & 3.4 & 1.01 & 2.85 & 1.34 \\\hline
Week 43 & 1.26 & 0.6 & 0.84 & 0.74 \\\hline
Week 44 & 3.99 & 0.59 & 0.79 & 1.3 \\\hline
Week 45 & 5.49 & 1.23 & 2.9 & 1.12 \\\hline
Week 46 & 2.16 & 0.87 & 1.43 & 0.85 \\\hline
Week 47 & 2.43 & 0.71 & 1.02 & 0.86 \\\hline
Week 48 & 4.09 & 0.63 & 1.05 & 1.04 \\\hline
Week 49 & 4.92 & 0.72 & 1.01 & 0.77 \\\hline
Week 50 & 1.42 & 0.83 & 1 & 0.77 \\\hline
Week 51 & 3.38 & 0.74 & 1.05 & 0.61 \\\hline
Week 52 & 6.21 & 0.65 & 1.1 & 0.82
\end{tabular}
\end{minipage}

\printbibliography

@article{tang_optimizing_2021,
	title = {Optimizing water and nitrogen managements for potato production in the agro-pastoral ecotone in North China},
	volume = {253},
	issn = {03783774},
	url = {https://linkinghub.elsevier.com/retrieve/pii/S0378377421002109},
	doi = {10.1016/j.agwat.2021.106945},
	pages = {106945},
	journaltitle = {Agricultural Water Management},
	shortjournal = {Agricultural Water Management},
	author = {Tang, Jianzhao and Xiao, Dengpan and Wang, Jing and Fang, Quanxiao and Zhang, Jun and Bai, Huizi},
	urldate = {2021-06-29},
	date = {2021-07},
	langid = {english},
}

@article{peake_effect_2018,
	title = {Effect of variable crop duration on grain yield of irrigated spring-wheat when flowering is synchronised},
	volume = {228},
	pages = {183--194},
	journaltitle = {Field Crops Research},
	author = {Peake, {AS} and Das, {BT} and Bell, {KL} and Gardner, M and Poole, N},
	date = {2018},
	note = {Publisher: Elsevier},
}

@article{collins_improving_2021,
	title = {Improving productivity of Australian wheat by adapting sowing date and genotype phenology to future climate},
	volume = {32},
	issn = {22120963},
	url = {https://linkinghub.elsevier.com/retrieve/pii/S2212096321000292},
	doi = {10.1016/j.crm.2021.100300},
	pages = {100300},
	journaltitle = {Climate Risk Management},
	shortjournal = {Climate Risk Management},
	author = {Collins, Brian and Chenu, Karine},
	urldate = {2021-07-30},
	date = {2021},
	langid = {english},
}

@article{rahimi-moghaddam_towards_2021,
	title = {Towards withholding irrigation regimes and drought-resistant genotypes as strategies to increase canola production in drought-prone environments: A modeling approach},
	volume = {243},
	issn = {03783774},
	url = {https://linkinghub.elsevier.com/retrieve/pii/S0378377420302729},
	doi = {10.1016/j.agwat.2020.106487},
	shorttitle = {Towards withholding irrigation regimes and drought-resistant genotypes as strategies to increase canola production in drought-prone environments},
	pages = {106487},
	journaltitle = {Agricultural Water Management},
	shortjournal = {Agricultural Water Management},
	author = {Rahimi-Moghaddam, Sajjad and Eyni-Nargeseh, Hamed and Ahmadi, Seyed Ahmad Kalantar and Azizi, Khosro},
	urldate = {2021-07-13},
	date = {2021-01},
	langid = {english},
}

@article{xiao_climate_2020,
	title = {Climate change impact on yields and water use of wheat and maize in the North China Plain under future climate change scenarios},
	volume = {238},
	issn = {03783774},
	url = {https://linkinghub.elsevier.com/retrieve/pii/S0378377420301888},
	doi = {10.1016/j.agwat.2020.106238},
	pages = {106238},
	journaltitle = {Agricultural Water Management},
	shortjournal = {Agricultural Water Management},
	author = {Xiao, Dengpan and Liu, De Li and Wang, Bin and Feng, Puyu and Bai, Huizi and Tang, Jianzhao},
	urldate = {2021-07-13},
	date = {2020-08},
	langid = {english},
}

@article{dutta_improved_2020,
	title = {Improved water management practices improve cropping system profitability and smallholder farmers’ incomes},
	volume = {242},
	issn = {03783774},
	url = {https://linkinghub.elsevier.com/retrieve/pii/S0378377420303796},
	doi = {10.1016/j.agwat.2020.106411},
	pages = {106411},
	journaltitle = {Agricultural Water Management},
	shortjournal = {Agricultural Water Management},
	author = {Dutta, S. K and Laing, Alison M. and Kumar, S. and Gathala, Mahesh K. and Singh, Ajoy K. and Gaydon, D.S. and Poulton, P.},
	urldate = {2021-07-13},
	date = {2020-12},
	langid = {english},
}

@article{ojeda_effects_2020,
	title = {Effects of soil-and climate data aggregation on simulated potato yield and irrigation water requirement},
	volume = {710},
	issn = {0048-9697},
	pages = {135589},
	journaltitle = {Science of The Total Environment},
	shortjournal = {Science of The Total Environment},
	author = {Ojeda, Jonathan J and Rezaei, Ehsan Eyshi and Remenyi, Tomas A and Webb, Mathew A and Webber, Heidi A and Kamali, Bahareh and Harris, Rebecca {MB} and Brown, Jaclyn N and Kidd, Darren B and Mohammed, Caroline L},
	date = {2020},
	note = {Publisher: Elsevier},
}

@article{schultz_can_2021,
	title = {Can deep learning beat numerical weather prediction?},
	volume = {379},
	issn = {1364-503X},
	pages = {20200097},
	number = {2194},
	journaltitle = {Philosophical Transactions of the Royal Society A},
	shortjournal = {Philosophical Transactions of the Royal Society A},
	author = {Schultz, Martin G and Betancourt, Clara and Gong, Bing and Kleinert, Felix and Langguth, Michael and Leufen, Lukas Hubert and Mozaffari, Amirpasha and Stadtler, Scarlet},
	date = {2021},
	note = {Publisher: The Royal Society Publishing},
}

@article{bauer_quiet_2015,
	title = {The quiet revolution of numerical weather prediction},
	volume = {525},
	issn = {1476-4687},
	url = {https://doi.org/10.1038/nature14956},
	doi = {10.1038/nature14956},
	pages = {47--55},
	number = {7567},
	journaltitle = {Nature},
	shortjournal = {Nature},
	author = {Bauer, Peter and Thorpe, Alan and Brunet, Gilbert},
	date = {2015-09-01},
}

@inproceedings{long_fully_2015,
	title = {Fully Convolutional Networks for Semantic Segmentation},
	pages = {3431--3440},
	booktitle = {Proceedings of the {IEEE} conference on computer vision and pattern recognition},
	author = {Long, Jonathan and Shelhamer, Evan and Darrell, Trevor},
	date = {2015},
}

@article{van_den_oord_wavenet_2016,
	title = {Wavenet: A generative model for raw audio},
	journaltitle = {{arXiv}:1609.03499},
	author = {van den Oord, Aaron and Dieleman, Sander and Zen, Heiga and Simonyan, Karen and Vinyals, Oriol and Graves, Alex and Kalchbrenner, Nal and Senior, Andrew and Kavukcuoglu, Koray},
	date = {2016},
}

@inproceedings{gehring_convolutional_2017,
	title = {Convolutional Sequence to Sequence Learning},
	volume = {70},
	url = {https://proceedings.mlr.press/v70/gehring17a.html},
	series = {Proceedings of Machine Learning Research},
	pages = {1243--1252},
	booktitle = {Proceedings of the 34th International Conference on Machine Learning},
	publisher = {{PMLR}},
	author = {Gehring, Jonas and Auli, Michael and Grangier, David and Yarats, Denis and Dauphin, Yann N.},
	editor = {Precup, Doina and Teh, Yee Whye},
	date = {2017-08-06},
}

@article{bai_empirical_2018,
	title = {An Empirical Evaluation of Generic Convolutional and Recurrent Networks for Sequence Modeling},
	journaltitle = {{arXiv}:1803.01271},
	author = {Bai, Shaojie and Kolter, J. Zico and Koltun, Vladlen},
	date = {2018},
}

@article{borovykh_conditional_2018,
	title = {Conditional Time Series Forecasting with Convolutional Neural Networks},
	journaltitle = {{arXiv}:1703.04691},
	author = {Borovykh, Anastasia and Bohte, Sander and Oosterlee, Cornelis W.},
	date = {2018},
}

@article{lilley_defining_2019,
	title = {Defining optimal sowing and flowering periods for canola in Australia},
	volume = {235},
	issn = {03784290},
	url = {https://linkinghub.elsevier.com/retrieve/pii/S0378429018320641},
	doi = {10.1016/j.fcr.2019.03.002},
	pages = {118--128},
	journaltitle = {Field Crops Research},
	shortjournal = {Field Crops Research},
	author = {Lilley, Julianne M. and Flohr, Bonnie M and Whish, Jeremy P M and Farre, Imma and Kirkegaard, John A},
	urldate = {2019-08-31},
	date = {2019-04},
	langid = {english},
}

@online{queensland_government_get_2024,
	title = {Get Point Data {\textbar} {LongPaddock} {\textbar} {SILO}},
	url = {https://www.longpaddock.qld.gov.au/silo/point-data/},
	author = {{Queensland Government}},
	date = {2024},
}

@online{noaa_national_oceanic_and_atmospheric_administration_what_2024,
	title = {What are El Niño and La Niña?},
	url = {https://oceanservice.noaa.gov/facts/ninonina.html},
	author = {{NOAA (National Oceanic and Atmospheric Administration)}},
	date = {2024},
}

@article{kingma_adam_2017,
	title = {Adam: A Method for Stochastic Optimization},
	journaltitle = {{arXiv}:1412.6980},
	author = {Kingma, Diederik P. and Ba, Jimmy},
	date = {2017},
}

@inproceedings{he_deep_2016,
	title = {Deep Residual Learning for Image Recognition},
	booktitle = {Proceedings of the {IEEE} Conference on Computer Vision and Pattern Recognition ({CVPR})},
	author = {He, Kaiming and Zhang, Xiangyu and Ren, Shaoqing and Sun, Jian},
	date = {2016-06},
}

@online{apsim_initiative_creating_2024,
	title = {Creating an {APSIM} met file using Excel},
	url = {https://www.apsim.info/support/apsim-training-manuals/creating-an-apsim-met-file-using-excel/},
	author = {{APSIM Initiative}},
	date = {2024},
}

@article{lorenz_deterministic_1963,
	title = {Deterministic nonperiodic flow},
	volume = {20},
	pages = {130--141},
	number = {2},
	journaltitle = {Journal of atmospheric sciences},
	author = {Lorenz, Edward N},
	date = {1963},
}

@article{hochman_re-inventing_2009,
	title = {Re-inventing model-based decision support with Australian dryland farmers. 4. Yield Prophet helps farmers monitor and manage crops in a variable climate},
	volume = {60},
	pages = {1057--1070},
	number = {11},
	journaltitle = {Crop and Pasture Science},
	author = {Hochman, Z and Van Rees, H and Carberry, PS and Hunt, JR and McCown, RL and Gartmann, A and Holzworth, D and Van Rees, S and Dalgliesh, NP and Long, W and {others}},
	date = {2009},
	note = {Publisher: {CSIRO} Publishing},
}

@article{ramirez-villegas_assessing_2017,
	title = {Assessing uncertainty and complexity in regional-scale crop model simulations},
	volume = {88},
	issn = {1161-0301},
	url = {https://www.sciencedirect.com/science/article/pii/S1161030115300666},
	doi = {10.1016/j.eja.2015.11.021},
	pages = {84--95},
	journaltitle = {Uncertainty in crop model predictions},
	shortjournal = {European Journal of Agronomy},
	author = {Ramirez-Villegas, Julian and Koehler, Ann-Kristin and Challinor, Andrew J.},
	date = {2017-08-01},
}

@article{fritz_comparison_2019,
	title = {A comparison of global agricultural monitoring systems and current gaps},
	volume = {168},
	issn = {0308-521X},
	url = {https://www.sciencedirect.com/science/article/pii/S0308521X17312027},
	doi = {10.1016/j.agsy.2018.05.010},
	pages = {258--272},
	journaltitle = {Agricultural Systems},
	shortjournal = {Agricultural Systems},
	author = {Fritz, Steffen and See, Linda and Bayas, Juan Carlos Laso and Waldner, François and Jacques, Damien and Becker-Reshef, Inbal and Whitcraft, Alyssa and Baruth, Bettina and Bonifacio, Rogerio and Crutchfield, Jim and Rembold, Felix and Rojas, Oscar and Schucknecht, Anne and Van der Velde, Marijn and Verdin, James and Wu, Bingfang and Yan, Nana and You, Liangzhi and Gilliams, Sven and Mücher, Sander and Tetrault, Robert and Moorthy, Inian and {McCallum}, Ian},
	date = {2019-01-01},
}

@article{dokoohaki_comprehensive_2021,
	title = {A comprehensive uncertainty quantification of large-scale process-based crop modeling frameworks},
	volume = {16},
	issn = {1748-9326},
	pages = {084010},
	number = {8},
	journaltitle = {Environmental Research Letters},
	shortjournal = {Environmental Research Letters},
	author = {Dokoohaki, Hamze and Kivi, Marissa S and Martinez-Feria, Rafael and Miguez, Fernando E and Hoogenboom, Gerrit},
	date = {2021},
	note = {Publisher: {IOP} Publishing},
}

@online{philosophical_transactions_of_the_royal_society_a_machine_2021,
	title = {Machine learning for weather and climate modelling},
	url = {https://royalsocietypublishing.org/toc/rsta/2021/379/2194},
	author = {{Philosophical Transactions of the Royal Society A}},
	date = {2021},
}

@inproceedings{van_den_oord_pixel_2016,
	location = {New York, New York, {USA}},
	title = {Pixel Recurrent Neural Networks},
	volume = {48},
	url = {https://proceedings.mlr.press/v48/oord16.html},
	series = {Proceedings of Machine Learning Research},
	pages = {1747--1756},
	booktitle = {Proceedings of The 33rd International Conference on Machine Learning},
	publisher = {{PMLR}},
	author = {van den Oord, Aäron and Kalchbrenner, Nal and Kavukcuoglu, Koray},
	editor = {Balcan, Maria Florina and Weinberger, Kilian Q.},
	date = {2016-06-20},
}

@article{pan_survey_2010,
	title = {A Survey on Transfer Learning},
	volume = {22},
	issn = {1558-2191},
	doi = {10.1109/TKDE.2009.191},
	pages = {1345--1359},
	number = {10},
	journaltitle = {{IEEE} Transactions on Knowledge and Data Engineering},
	shortjournal = {{IEEE} Transactions on Knowledge and Data Engineering},
	author = {Pan, Sinno Jialin and Yang, Qiang},
	date = {2010-10},
}

@article{bihlo_generative_2021,
	title = {A generative adversarial network approach to (ensemble) weather prediction},
	volume = {139},
	issn = {0893-6080},
	url = {https://www.sciencedirect.com/science/article/pii/S0893608021000459},
	doi = {10.1016/j.neunet.2021.02.003},
	pages = {1--16},
	journaltitle = {Neural Networks},
	shortjournal = {Neural Networks},
	author = {Bihlo, Alex},
	date = {2021-07-01},
}

\end{document}